\documentclass[twocolumn,
superscriptaddress,
 amsmath,amssymb,
 aps,pra,showkeys
]{revtex4-1}

\usepackage{color}
\usepackage{subfig}

\usepackage{graphicx}

\AtBeginDocument{
  }

\begin{document}


\title{Level Generation with Quantum Reservoir Computing}

\author{Jo\~ao S. Ferreira}
\email{joao@mothquantum.com}
\affiliation{Moth, Arlesheim, BL, Switzerland}

\author{Pierre Fromholz}

\affiliation{Moth, Arlesheim, BL, Switzerland}

\author{Hari Shaji}
\affiliation{Moth, Somerset House, London, United Kingdom}

\author{James R. Wootton}
\affiliation{Moth, Arlesheim, BL, Switzerland}

\begin{abstract}
Reservoir computing is a form of machine learning particularly suited for time series analysis, including forecasting predictions. We take an implementation of \emph{quantum} reservoir computing that was initially designed to generate variants of musical scores and adapt it to create levels of Super Mario Bros. Motivated by our analysis of these levels, we develop a new Roblox \textit{obby} where the courses can be generated in real time on superconducting qubit hardware, and investigate some of the constraints placed by such real-time generation.
\end{abstract}


\maketitle

\section{Introduction}
After many years of development, quantum computing hardware is rapidly developing towards commercialization. This has led to an explosion of interest in quantum algorithms and applications~\cite{Abbas2024}. In the games industry the intersection of quantum computing and games has been explored for almost a decade~\cite{piispanen2024definingquantumgames}. Although most examples are currently within an educational context~\cite{Seskir2022}, the potential of quantum computing for procedural generation has started to be  explored~\cite{cog2020,quantumblur,olart,karamlou-etal-2022-quantum}. Here we go beyond this previous proof-of-principle work by developing an example of quantum procedural generation that can provide real-time generation of levels within a live game.

Specifically, we explore a generative system based on Quantum Reservoir Computing (QRC)~\cite{Mujal2021} to generate game levels. QRC is a type of machine learning algorithm that is well-suited to operate within the constraints of currently existing quantum computers called Noisy Intermediate-Scale Quantum (NISQ) computers~\cite{Preskill2018quantumcomputingin}, making it a practical choice for near-term applications.
Indeed, we estimate that the games discussed in this paper require 10 to 20 QRC processes running in parallel without quantum communication between them and each using 6 or 7 qubits (for a total of 60 to 140 qubits). Such requirements are well within the capabilities of many current quantum devices, such as the latest hardware of IBM Quantum, Google Quantum AI and IQM. 

QRC uses significantly fewer trainable parameters compared to other generative methods used in games such as generative adversarial networks~\cite{gans} or variational autoencoders~\cite{vae}.
The potential of using small datasets is desirable, since it means that one can train only on one's own content, without the need for large datasets with possible copyright infringement. Specifically, we consider the case where the process is trained on a single level of Super Mario Bros or a Roblox \emph{obby}, and then used to produce variants of that level.
We then investigate how distinct the generations are from the original, including novel segments, while also maintaining gameplay integrity. To evaluate these two performances, we use two metrics that we define, the \emph{originality} and the \emph{rate of broken sequences}, that allow us to find the optimal hyper parameters for the generation. For practical reasons, we resort to simulating the quantum circuit on non-quantum computers and leave the training and generation on real quantum hardware for later works. 

We present the specifics of our QRC implementations in Sec.~\ref{sec:QRC} and apply it to generate Super Mario Bros levels in Sec.~\ref{sec:SMB}. In Sec.~\ref{sec:Roblox} we discuss our Roblox \emph{obby} and conclude in Sec.~\ref{sec:conclusions} with the advantages and current limitations of using QRC for procedural generations in games.

\section{Quantum Reservoir Computing}\label{sec:QRC}

\begin{figure}[t]
  \centering

  \subfloat[Diagram of the data pipeline. The feedback loop for the generation mode is illustrated using dashed arrows.]{
    \includegraphics[width=\columnwidth]{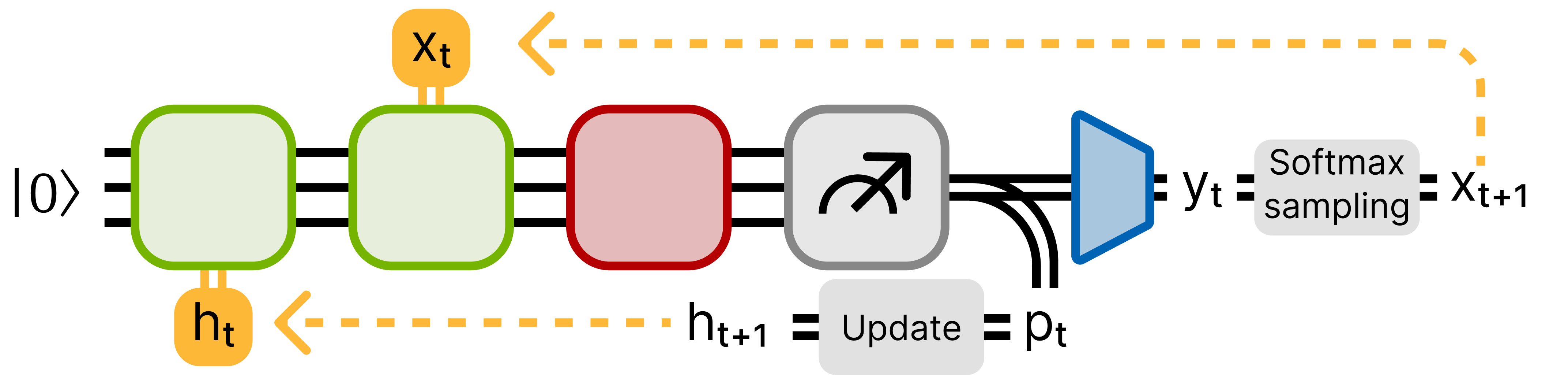}
    \label{fig:qrc-all}
  }\\[1ex]

  \subfloat[Left: example of a random circuit composed of randomly sampled X, H or CNOT gates. Right: embedding circuit with classical inputs encoded as $R_y$ rotations interlaced with CNOT gates. Both the memory $h_t$ and the input $x_t$ states uses this circuit.]{
    \begin{minipage}{.49\columnwidth}
      \centering
      \includegraphics[width=\linewidth]{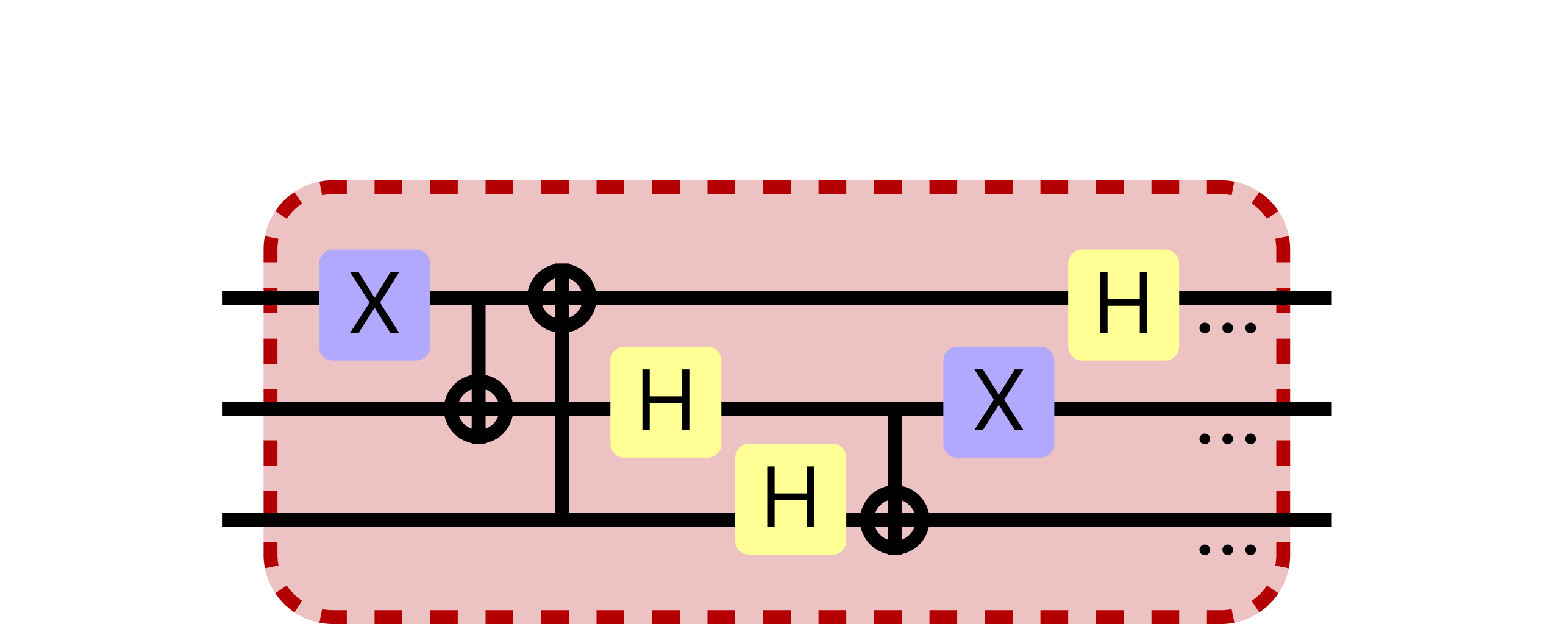}
    \end{minipage}\hfill
    \begin{minipage}{.49\columnwidth}
      \centering
      \includegraphics[width=\linewidth]{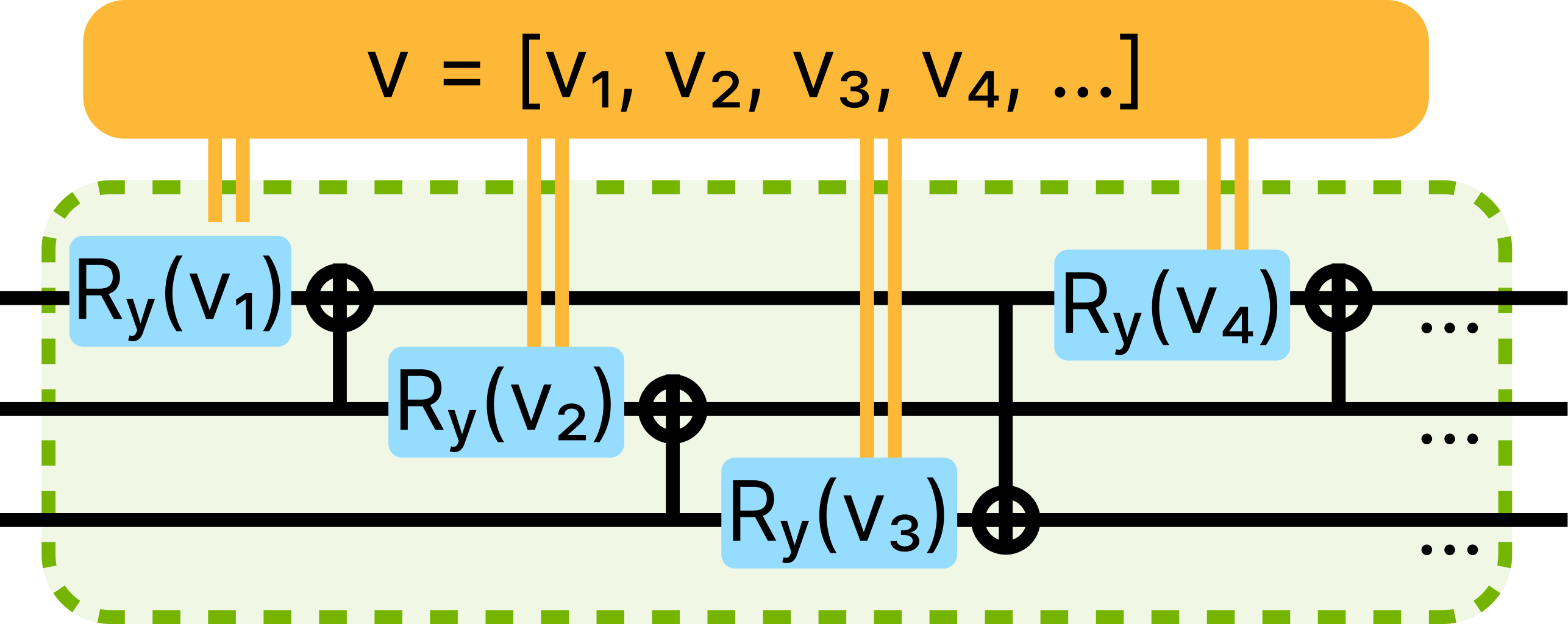}
    \end{minipage}
  }
  \caption{Diagram and quantum circuits of the quantum reservoir computing algorithm used in this paper. The input $x_t$ at step $t$ is encoded as parameters of a complex quantum circuit whose measurement outcome is fed to a FNN to sample the next feature.}
  \label{fig:QRC}
\end{figure}

Reservoir computing is a form of machine learning in which the majority of information processing is performed by a reservoir: a physical system whose complex and sometimes chaotic dynamics allows it to explore the available state space more efficiently than a systematic and deterministic algorithm~\cite{Choi2020}. This reservoir is supplemented by a trainable output layer, which converts the outputs of the reservoir into the outputs of the desired computation. 

Quantum Reservoir Computing (QRC) is an implementation of a reservoir computing architecture where the reservoir consists of a set of entangled qubits. When used for generation tasks, QRC belongs to the class of recurrent algorithms that, given a sequence of inputs $\{x_0, \dots, x_{t-1}\}$, predicts the next value $x_t$ leveraging a hidden memory state. 
As in any reservoir computing approach, the input signal is injected into the reservoir, which evolves under a complex, random dynamic that supports temporal correlations. The resulting quantum state is measured and passed through a neural network to produce the output signal. A key characteristic is that the reservoir's dynamics remain fixed, and only the neural network is trainable. This significantly reduces the number of parameters and eliminates the need for back-propagation, making the approach well-suited for current NISQ-era quantum devices.

In this work, we focus on a \textit{hybrid} implementation of QRC~\cite{ISQCMC_miranda}, originally developed for music generation~\cite{qrc-music}; see Fig.~\ref{fig:QRC}a for a schematic diagram. To use it for level generation, we must first decompose each level sequentially into $f$ distinct features corresponding to the inputs $x_t\in \{0,...,f-1\}$.
QRC consists of a quantum reservoir composed of $q$ qubits, partitioned into two parametric circuits that encode the input $x_t$ and hidden state $h_t$ and a third random circuit, along with a classical linear feed-forward neural network (FNN) responsible for predicting the output vector $y_t$ of dimension $f$. The hidden state is purely classical and is obtained by measuring the quantum probability vector $p_t$.

QRC operates in two modes: training and generation. During training, the predicted output $y_t$ is compared to the target vector $y_t^{\text{target}}$ from the training data~\footnote{$y_t^{\text{target}}$ is the one-hot representation of $x_{t+1}$ and is thus a vector of length $f$}, and the FNN parameters are updated to minimize a cross-entropy loss function. In the generation phase, the outputs $p_t$ and $y_t$ at time step $t$ become the inputs for time step $t+1$ according to the update rule:
\begin{align}
    x_{t+1} &= \text{Categorical}[\text{softmax}(y_t/T)], \\
    h_{t+1} &= (1 - \epsilon) h_t + \epsilon p_t,
\end{align}
where $\epsilon$ is the leaking rate 
(set to 0.3 in this work) and $T$ is a controllable temperature parameter that regulates the variability of predictions. Because $h_t$ iteratively encodes past predictions, it is often referred to as the \textit{memory} of the model. As both this state and the prediction are stored in a buffer to prepare the next step of the algorithm, the version of QRC we are using is called a feedback controlled QRC
protocol~\cite{murauer2025}.

We encode both the hidden and input states as angles for $R_y$ rotation gates, interlaced by CNOT gates (see Fig.~\ref{fig:QRC}b, right). This design ensures that, for each physically accessible quantum outcome, there exists at least one corresponding input encoding. This property is referred to as \textit{universality} and is considered essential for an effective reservoir~\cite{monzani2024}.  Because large-angle rotations are inherently non-linear, even small variations in the input can cause significant changes in the quantum state which effectively contributes to the chaotic property desired in reservoir systems.
Additionally, the quantum circuit includes a fixed, randomly sampled sequence of gates from the set $\{X, H, \text{CNOT}\}$ (see Fig.~\ref{fig:QRC}b, left), further enhancing the chaotic nature of the reservoir dynamics.

A brief note is warranted on the measurement process: due to shot limitations (4000 shots in our experiments), measurements are inherently stochastic and non-invertible. Although this could, in principle, degrade the performance of the generation, previous work suggests that a small degree of stochasticity allows $p_t$ to explore a broader portion of the Hilbert space, which could lead to better results~\cite{domingo2023takingadvantagenoisequantum}. The same argument applies to the presence of noise, whether from simulated or real quantum hardware. Determining the impact of noise is a question we explore later in the paper.

In the remainder of this work, we treat the QRC protocol as a black box and focus on tuning its hyperparameters, such as the number of qubits $q$ and the temperature $T$.

\section{Application to Super Mario Bros} \label{sec:SMB}

Levels of Super Mario Bros are ideal test cases for level generation algorithms, since they are both simple and familiar to many. They therefore serve as the perfect first benchmark for level generation using QRC.

\begin{figure*}[t]
  \centering
  \subfloat[Original level 1-2 composed of 32 distinct features.]{
    \includegraphics[width=\textwidth]{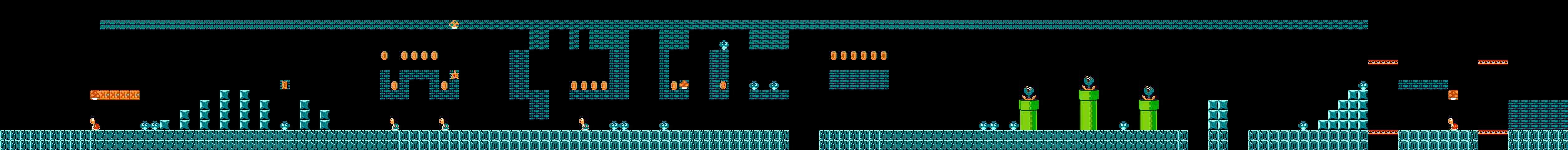}
    \label{fig:level-original}
  }\\[1ex]
  \subfloat[QRC-generated level with the optimal temperature $T=1$.]{
    \includegraphics[width=\textwidth]{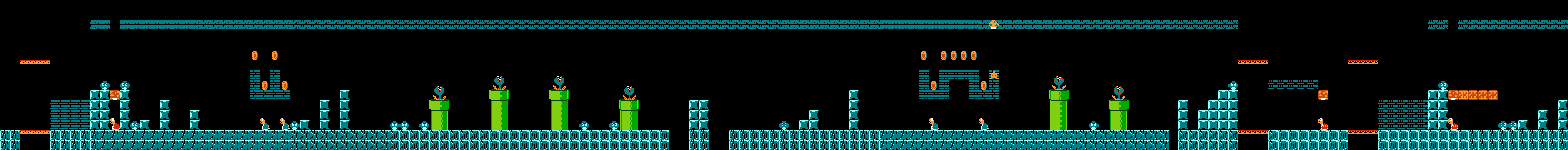}
    \label{fig:level-T1}
  }\\[1ex]
  \subfloat[Half of a QRC-generated level with $T=0.01$ (left) and $T=10$ (right).]{
    \begin{minipage}{.48\textwidth}
      \includegraphics[width=\linewidth]{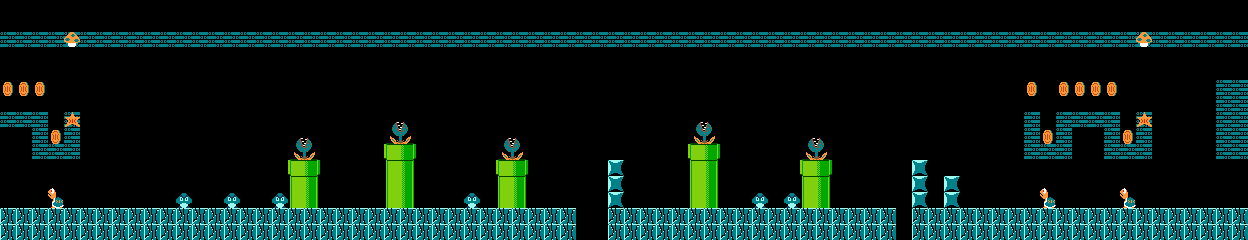}
    \end{minipage}\hspace{.03\textwidth}
    \begin{minipage}{.48\textwidth}
      \includegraphics[width=\linewidth]{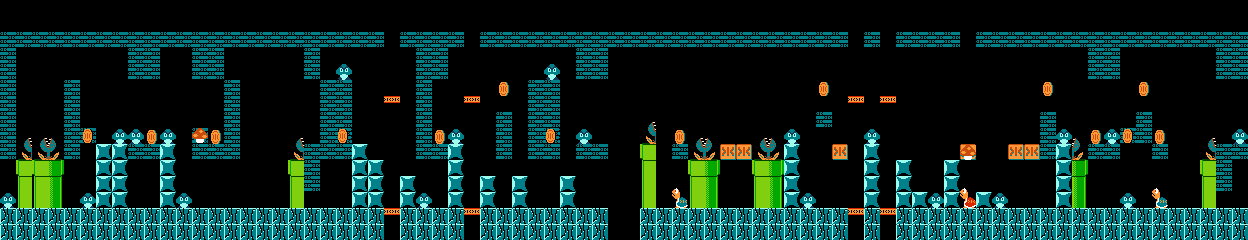}
    \end{minipage}
  }
  \caption{The original level, with a selection of QRC-generated variants.}
  \label{fig:levels}
\end{figure*}

To perform this level generation, we must first express levels as a sequence of features. We do this by interpreting each level as a sequence of columns, each 16 pixels wide (the width of a block). All the unique columns that occur in the level are then labelled by numbers and used as features. In the original first four levels, there are a large number of unique features in each level. However, the differences in many cases are due to minor variations, such as background art, which do not impact gameplay. By filtering out these elements to focus on gameplay aspects of the generations only, each level can be expressed using around 30 unique features. 
In this paper, we exclusively use the level ``1-2'' (see Fig.\ref{fig:levels}a) with a size of 157 features composed of 32 unique features. For the purposes of our study, we refer to this as the \textit{original level} and use it as training data. 

\subsection{Generation on ideal hardware}

We first train our QRC system on a simulation of ideal quantum hardware, without the effects of noise. For now, we fix the number of qubits to 6, vary the temperature parameter $T \in [0.01, 30]$ and generate 100 variants, each matching the length of the original level. 

Determining what makes a \textit{good} level is not straightforward~\cite{bazzaz2025level}, but we consider three key objectives that a generated level must balance. First, it should introduce \textit{transitions} (sequences of two features) that were not present in the original level, thus ensuring novelty. Second, it should minimize sequences that break gameplay, such as the left half of a pipe not connecting immediately to the right half, henceforth named \textit{broken} sequences or \textit{broken} transitions. Third, the generated level should preserve the large-scale structures of the original level, ensuring that it remains familiar.

Our initial experiments revealed distinct trends in the generated levels depending on the temperature parameter.
\begin{itemize}
    \item At low temperatures ($T \lesssim 1$), the levels displayed long repetitions of the original features, with sequences extending up to 20 features in length. These levels contained very few broken transitions but lacked the variability required for meaningful novelty.
    \item On the other hand, at high temperatures ($T \gtrsim 1$), the levels became highly random, characterized by very short feature sequences and a large number of broken sequences.
    \item The most balanced results were observed for $T \sim 1$, where the generated levels achieved a trade-off between novelty, structural coherence, and minimal broken sequences. In this regime, the levels were mostly distinct from the original, yet certain recognizable sequences such as the coin-filled room occasionally reappeared, suggesting that the large-scale structure was being partially preserved.
\end{itemize}
 
To further analyse the performance of the QRC, we compared its output to levels generated by two alternative benchmark generative algorithms. The first is uncorrelated sampling and involves generating features based on their frequency in the original level, without considering any relationships between neighbouring features. The second is a Markov chain model, which captures the pairwise transitions between features in the original level and generates sequences based on these observed transitions.

Note that the task we consider, taking existing content and riffing on it, is also performed by the popular \textit{Wavefunction collapse} algorithm\cite{Gumin_Wave_Function_Collapse_2016}. However, for the problem of generating 1D sequences as considered here, WFC essentially reduces to a Markov chain. Comparisons to WFC and will then become relevant in future work, as we generalize our QRC approach to 2D content.

\begin{figure}[t]
  \includegraphics[width=0.45\textwidth]{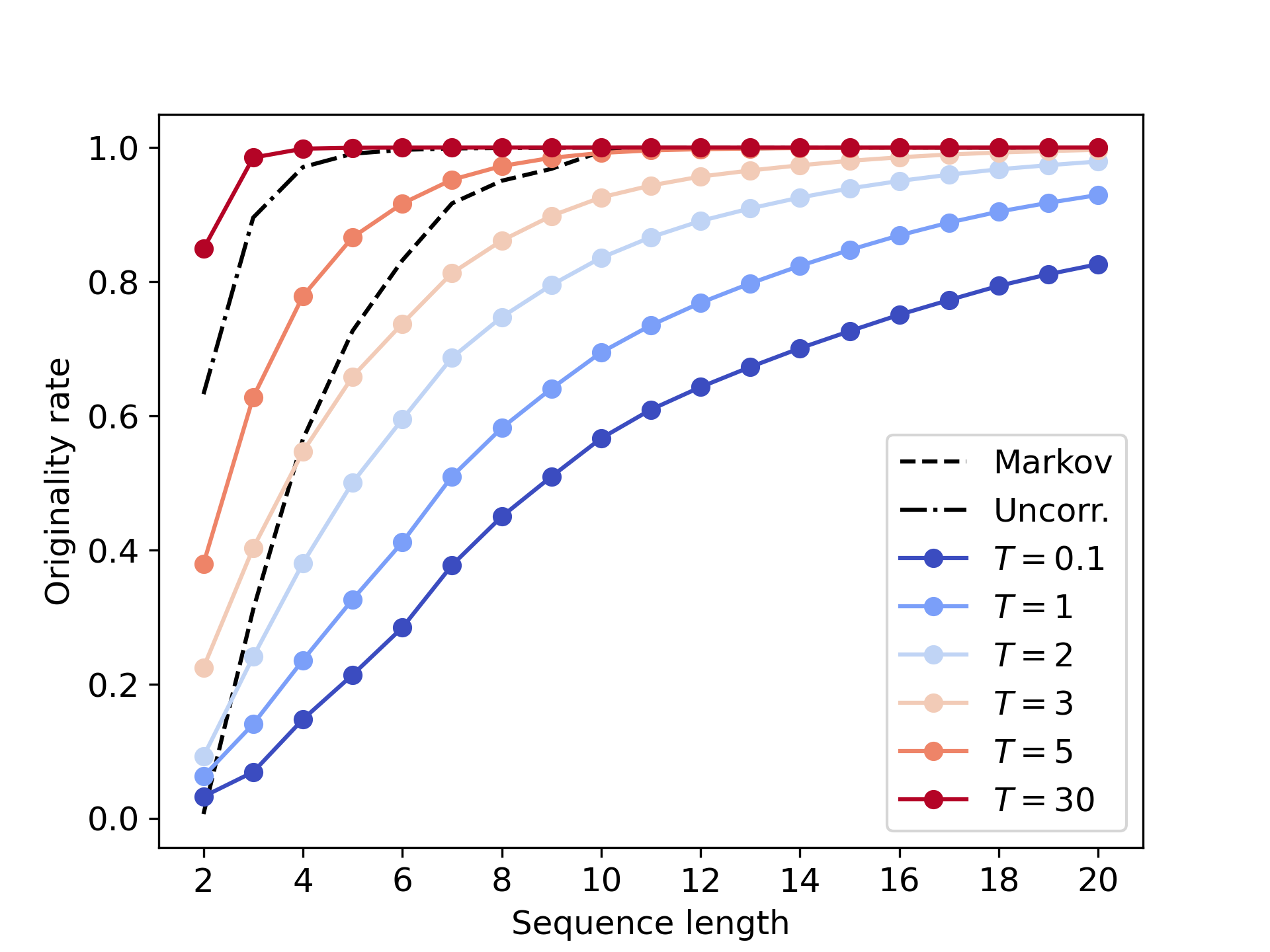}
  \caption{The rate of new sequences of a certain length for the uncorrelated generator (dotted dashed line), the Markov chain generator (dashed line) and QRC generator for different values of the temperature $T$ (dots). An ideal generator will have an originality rate higher than the Markov chain for short sequences but lower for long sequences, ensuring a balance of novelty and familiarity. \label{fig:originality}}
\end{figure}

To assess the originality of the generated levels, we introduce in Fig.~\ref{fig:originality} the \textit{originality rate} defined as the fraction of the generated sequences of length $L$ that are not present in the original level, normalized by the total number of sequences of length $L$ in the original levels. An ideal level would have a mostly flat profile, being capable of generating new transitions while keeping the originality at large-scale structures ($L\sim10$) smaller than 1.

As expected, the uncorrelated generator is inherently incapable of replicating most sequences of any length from the original level, providing a baseline for originality (otherwise upper bounded by 1).

By design, the Markov chain generator always reproduces the original transitions, never making a broken transition, but also never proposing anything new. It also quickly saturates for sequences larger than 10 features. It follows that the Markov chain struggles to replicate the larger-scale structures present in the original level.

\begin{figure}[t]
  \includegraphics[width=0.45\textwidth]{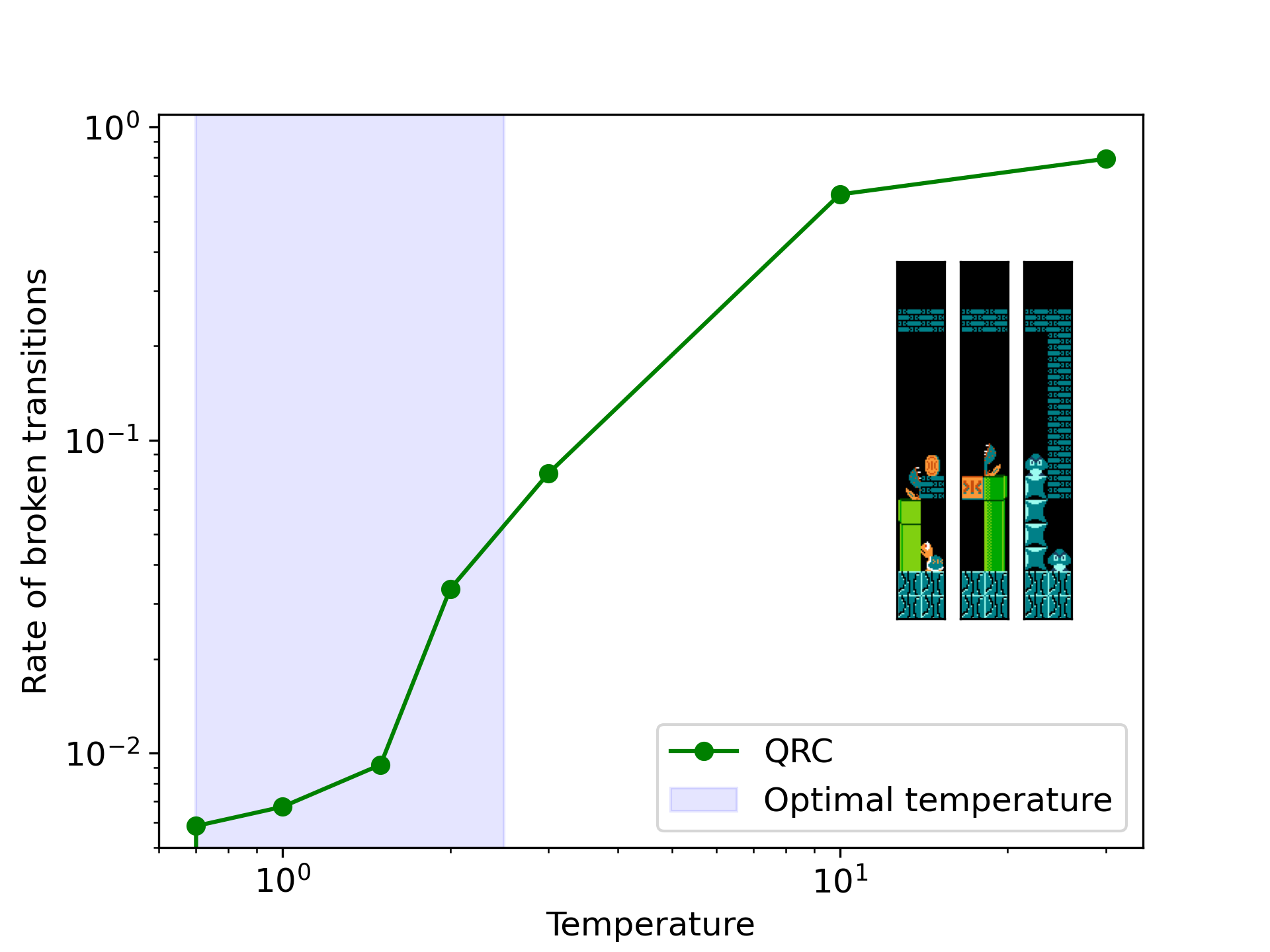}
  \caption{Rate of transitions that break gameplay. In blue we highlight the region in which the temperature gives the best result, upper bounded by the condition that the error rate remains relatively low (below 5\%), and lower bounded by the restriction that the levels are not found to be too repetitive. Inset: examples of three broken transitions. \label{fig:error}}
\end{figure}

The QRC, particularly for temperatures around $T = 1$, demonstrates a good balance in this regard. The slower growth means that sequences of length 15 occasionally appear, preserving recognizable features that make the generated levels identifiable. Simultaneously, the QRC can surpass the Markov chain in originality for shorter sequences (e.g., $L=2,3$) thus introducing novel transitions. However, this increase in originality naturally leads to a higher rate of broken transitions shown in Fig.~\ref{fig:error}. The error rate is defined as the number of broken transitions divided by the total number of transitions involving breakable features (e.g., the two halves of a pipe). Notably, the error rate remains low (below 5\%) for temperatures as high as $T=2$ where the originality for sequences of length 2 and 3 exceeds and matches, respectively, that of the Markov generator.

An important advantage of this approach is that the temperature parameter $T$ can be adjusted \textit{post-computation}, meaning that a game developer can easily tune the originality-to-error ratio without incurring additional computational overhead. Based on our analysis, we recommend restricting the temperature to the range highlighted in the blue regions of Fig.~\ref{fig:error}, where the error rate remains comparable to that of the Markov baseline while achieving higher originality and better preservation of large-scale structures.

\subsection{Generation on noisy hardware}

So far, our results are based on noiseless quantum simulations, ensuring optimal distinguishability between circuit outputs and, consequently, the best possible performance during training and generation. However, as the number of qubits increases, full simulation becomes computationally infeasible—typically beyond 30 qubits for general-purpose circuits. Beyond this point, training and generation must be performed on real quantum hardware, where the impact of noise during both training and generation—becomes a critical concern~\cite{fry2023optimizing,yasuda2023quantumreservoircomputingrepeated}.

One obvious way to do this is to run directly on current quantum harwdare. However, due to the large number of gates and shots required in our QRC algorithm, training the model on real quantum hardware is prohibitively slow and expensive. Moreover, the system’s behavior fluctuates over time due to so-called noise drift, making it difficult to reuse a trained model reliably at a later stage. For these reasons, the development and discussion of a full hardware implementation is beyond the scope of this work.

Instead, we assess the effects of noise using two different methods to simulate real noisy computers~\cite{Ding2020}: 
\begin{enumerate}
    \item Simulation with depolarizing noise applied to all 2-qubits gates with probability $p$ and 1-qubit gates with probability $p/10$;
    \item Simulation using a realistic noise model based on the calibration data from IQM's \emph{Garnet} device.
\end{enumerate}
The noise is present both for the training and generation, offering insight into the end-to-end performance one might expect on real quantum hardware.

The depolarizing noise model is the simplest: results are exact when $p = 0$, and become uniformly random as $p \to 1$ (maximal noise on two qubit gates). Current quantum hardware typically operates with $p \approx 3\%$~\cite{Liepelt_2024}. The second model incorporates more specific noise mechanisms tailored to the physics qubit systems, accounting for the so-called $T1$, $T2$, and read-out fidelities for each qubits, gate fidelities for each gate, and accounts for the geometry of the hardware as well as the specificities of the transpilation. Specifically, we use the daily updated calibration from IQM’s \emph{Garnet} processor in this paper.

\begin{figure}[t]
  \includegraphics[width=0.45\textwidth]{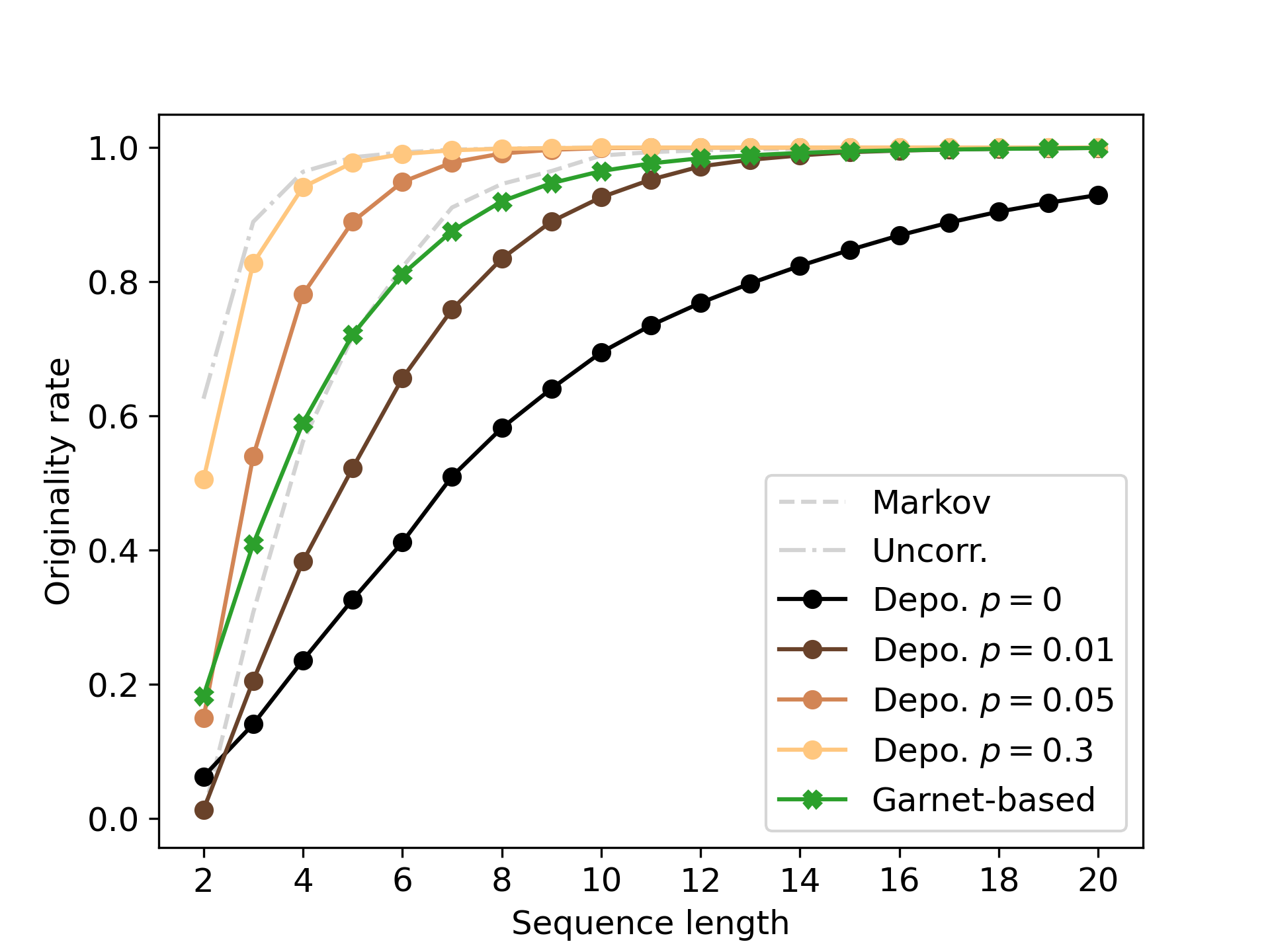}
  \caption{The originality metric when training and generating under different noise models. The depolarizing noise (dots) has the controlled depolarization rate $p$ with typical quantum hardware operating around 3\% depolarization rate. The noise model based on IQM's \emph{Garnet} process (crosses) is comparable to Markov and compatible with the specified mean polarization of 2\%-3\%.\label{fig:originality_noise}}
\end{figure}

In Fig.~\ref{fig:originality_noise}, we present the originality metric across different noise models at a fixed temperature of $T=1$. Depolarizing noise (dots) behaves similarly to an effective temperature increase, though it leads to a steeper rise in originality. Even a small depolarization rate of 1\% significantly diminishes the QRC’s ability to replicate sequences longer than 15 features. It appears that the theoretical advantage in adding noise~\cite{domingo2023takingadvantagenoisequantum} does not translate to better level generation. 

Nevertheless, there is a measurable improvement with respect to Markov for polarizations below 2\%, close to the Garnet-based noise model (crosses). The Garnet-based curve crosses the Markov (dashed line) for sequences of 6 features, suggesting that further optimization in circuit design and readout might showcase a more solid advantage when running on actual quantum devices. In-depth discussion of this topic is left for future work.

\section{Application to Roblox}\label{sec:Roblox}

Next, we consider an obstacle course (or \textit{obby}) implemented in Roblox. 
Our objective is to perform level generation via QRC, ensuring that the process can be executed on real quantum hardware as well as just emulations thereof. This generation will occur in real time, meaning that the time required to generate a new course must match the time expected to be taken to play the previous one. For reference, prior implementations of QRC using 32 features on superconducting qubits demonstrated a generation time of approximately 10 seconds to generate each element in a sequence. However, an experienced player may spend significantly less time, potentially as little as 1 second—per feature during gameplay. To ensure that course generation does not exceed the minimum play time, at least 10 levels must be generated in parallel.

This requirement places stringent constraints on the number of qubits that can be used in the QRC. In this section, we investigate how the number of qubits impacts the quality of the generated levels, with a particular focus on balancing quantum resource limitations with the demands of real-time gameplay.
\subsection{Analysis of Generated Content} \label{sec:Roblox_analysis}

\begin{figure}[t]
  \centering

  \subfloat[Small extract of the original level with the sequence \eqref{eq:seq} highlighted.
    \label{fig:roblox-original}]{
    \includegraphics[width=0.95\columnwidth]{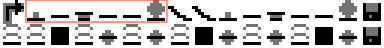}
  }\\[1ex]

  \subfloat[Extract of a QRC-generated level with 7 qubits capable of replicating \eqref{eq:seq}.
    \label{fig:roblox-qrc}]{
    \includegraphics[width=0.95\columnwidth]{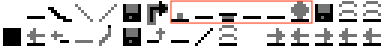}
  }\\[1ex]

  \subfloat[The first 6 features of (b) shown in the actual Roblox game.
    \label{fig:roblox-real}]{
    \includegraphics[width=0.95\columnwidth]{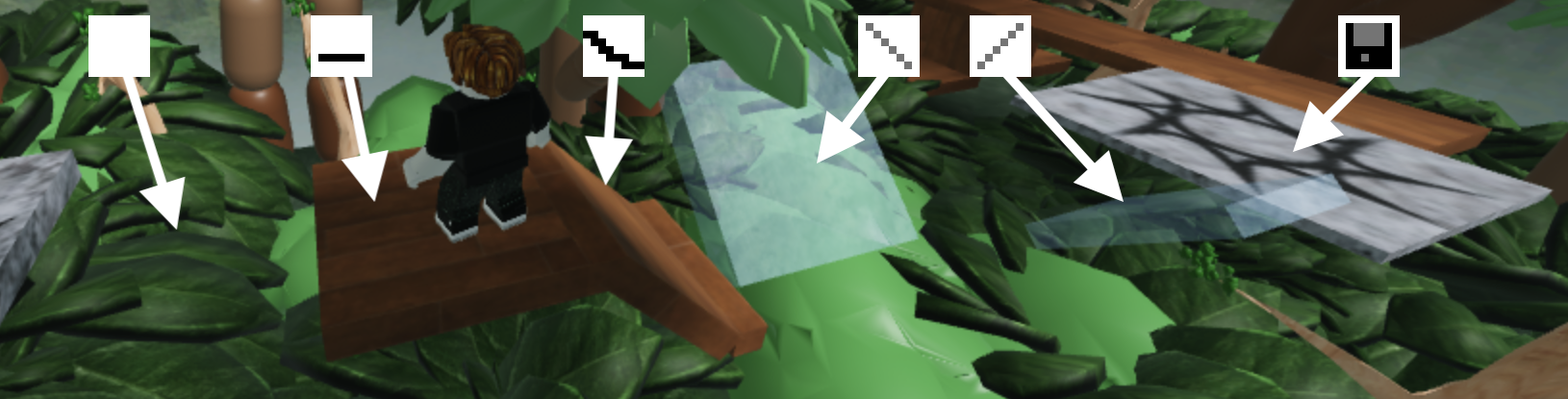}
  }

  \caption{Extracts of the original and generated levels before encoding and in the final game.}
  \label{fig:roblox-all}
\end{figure}

Similar to the levels in Super Mario Bros, each course in Roblox will be constructed from a set of 32 distinct features. These include spawn/save points, platforms, tilted platforms, rotating platforms, gaps, enemies, and other obstacles. The hand-designed course used to train the generator in this section is partially shown in Fig.~\ref{fig:roblox-original}. We designed the level such that in between each saving point only 6 distinct features are present. This means that we must consider longer levels of 288 features to ensure enough variability in the transitions and avoid over-fitting the FNN.

\begin{figure}[t]
  \includegraphics[width=0.45\textwidth]{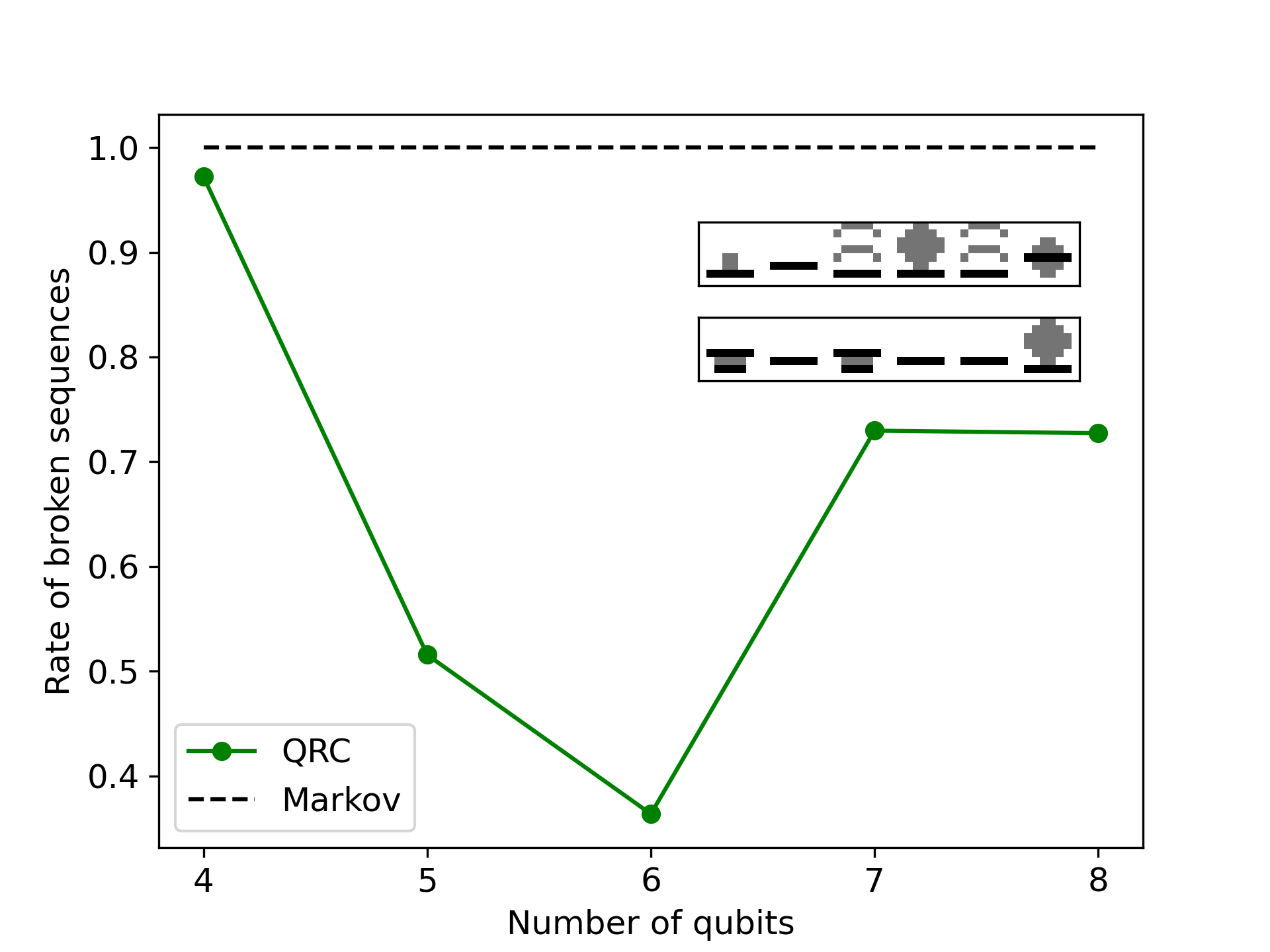}
  \caption{Fraction of broken sequences that include either a button, ball or hiding feature. Inset: two examples of such broken sequences. For Roblox level generations, 7 qubits is optimal. \label{fig:roblox-error}} 
\end{figure}

In this setup, each feature connects directly to the next, ensuring that broken transitions are minimized by construction. However, to further analyze the generator's ability to avoid critical errors, we introduce a long sequence which has the potential to produce game-breaking issues if not reproduced successfully:
\begin{equation}
    \includegraphics[height=5mm]{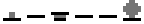},   \label{eq:seq}
\end{equation}
consisting of the following elements: 
\begin{itemize}
    \item A button (\includegraphics[height=\fontcharht\font`\B ]{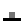}) that activates a rolling ball (\includegraphics[height=\fontcharht\font`\B ]{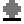}),
    \item A series of basic platforms (\includegraphics[height=\fontcharht\font`\B ]{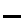}) that must be traversed while avoiding the ball, and 
    \item A hiding feature (\includegraphics[height=\fontcharht\font`\B ]{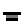}) that provides safety after jumping off the platforms.
\end{itemize}
If this sequence is not generated in the correct order ($\includegraphics[height=\fontcharht\font`\B ]{figs/single_button.png} \to \includegraphics[height=\fontcharht\font`\B ]{figs/single_hidding.png} \to \includegraphics[height=\fontcharht\font`\B ]{figs/single_ball.png}$), the game becomes unplayable. The associated error rate, defined as the fraction of broken sequences that include a button, ball, or hiding feature, is shown in Fig.~\ref{fig:roblox-error} as a function of the number of qubits used, $q$.
Our results indicate that QRC significantly outperforms the Markov chain generator, achieving an error rate as low as 36\% for the optimal number of 6 qubits. 

Another important characteristic of the generated levels is the frequency of spawn/save points. In the original hand-designed level, a save point appears every 16 features. The frequencies generated by different methods are summarized in the table below, which compares QRC, the Markov generator, and the uncorrelated generator.

\begin{center}
\begin{tabular}{c || c }
             & Separation between save points \\ [0.5ex] 
 \hline \hline
Original     & $16.0 \pm 0.0$     \\
Uncorrelated & $16.0  \pm 5.8$   \\
Markov       & $16.4  \pm  5.5$ \\
QRC $q=4$ & $17.9  \pm 7.9$    \\
QRC $q=5$ & $16.3  \pm 2.9$    \\
QRC $q=6$ & $18.8  \pm 4.1$   \\
QRC $q=7$ & $18.6  \pm 3.5$    \\
QRC $q=8$ & $17.1  \pm 4.1$           
\end{tabular}\label{tab:save_point_frequency}
\end{center}

The QRC with 6 qubits is capable of generating regular saving points with comparable mean separation to the original level, while achieving a reduced standard deviation compared to the Markov generator. This demonstrates QRC's ability to produce levels that are both structured and consistent. 

These results underscore the importance of selecting an appropriate number of qubits, which must be carefully tailored to the complexity of the training level. If too few qubits are used, the number of representable features becomes saturated, increasing the rate of broken sequences. Conversely, using too many qubits risks overfitting the FNN, particularly if the training data lacks sufficient variability. This overfitting can lead to a diminished capacity to replicate long sequences. Moreover, more qubits require more shots to sample the probability vector $h_t$, increasing the computation time.

\section{Conclusions\label{sec:conclusions}}

In this work, we explored the application of quantum reservoir computing (QRC) to generate sequential levels of video games, focusing on two case studies: Super Mario Bros and a custom-made Roblox obstacle course (\emph{obby}). By carefully selecting the key parameters, such as the temperature and the number of qubits, we demonstrated that QRC is capable of generating high-quality levels. In some metrics, such as original transitions and preservation of structural features, QRC was found to surpass a simple Markov generator. 

One of the key advantages of QRC lies in its tunability, allowing it to easily adapt to specific game requirements. Developers can fine-tune the originality/repetition ratio in real-time, offering significant flexibility in level design without requiring additional computational overhead. This makes QRC a powerful and practical tool for procedural content generation. The drawback of this approach is the increase of game-breaking transitions and sequences that will need to be corrected \emph{a posteriori}. 

In future work, we aim to implement the QRC algorithm on real quantum hardware, specifically using architectures such as IQM's \emph{Garnet}. Since training would primarily rely on simulations, we anticipate challenges related to domain alignment between simulated and physical systems. Potential mitigation strategies include, but are not limited to, optimizing circuit design to reduce noise sensitivity, incorporating non-linear readout layers, and training models using more sophisticated noise simulations.

Finally, we encourage readers to explore the Roblox game we developed (see Ref~\cite{nerii_game}) and evaluate for themselves the quality and playability of the generated levels. Player feedback will be invaluable in refining the QRC framework further and understanding its practical potential in real-world gaming applications.

\section*{Acknowledgments}
We thank Spencer Topel, Shreyan Chowdury, and Daniel Bultrini for their valuable suggestions when developing the algorithm and their helpful feedback during the writing of this paper.

\section*{Data availability}

All the data presented in this paper is available for free in Moth's Open Data repository~\cite{mothdata}.

\bibliographystyle{apsrev4-1}
\bibliography{references}

\end{document}